\documentclass[]{style}
\usepackage[most]{tcolorbox}        % tcolorbox高级特性（加载完整库）
\usepackage{amsmath,amssymb}
\usepackage{lipsum}                 % 假文生成
\usepackage{xcolor}                 % 颜色
\usepackage{fancyhdr}               % 页眉（仅保留空样式）
\usepackage{booktabs}
\usepackage[table]{xcolor}
\usepackage{url}
\usepackage{graphicx}   % 用于插入图片
\usepackage{subcaption} % 用于生成子图和子标题
\usepackage{multirow}
\usepackage{makecell}
\usepackage{bm}
\definecolor{lightpurple}{RGB}{153, 102, 204}  % 马卡龙浅紫（好看且清晰）
\definecolor{lilac}{RGB}{182, 133, 210}        % 丁香紫（更柔和）
\usepackage{hyperref}
\hypersetup{
    colorlinks=true,      % 启用彩色链接，而不是给链接加方框
    linkcolor=purple,        % 内部交叉引用的颜色（如图表、公式）
    citecolor=lilac,       % 参考文献引用的颜色（这里设为蓝色）
    urlcolor=magenta      % 网页 URL 的颜色
}
\usepackage{tabularx}

\definecolor{pwYellow}{RGB}{252,238,210}
\definecolor{pwPurple}{RGB}{239,220,247}
\definecolor{pwBlue}{RGB}{218,233,252}
\definecolor{pwGreen}{RGB}{218,241,234}

\definecolor{abstractpurple}{HTML}{9C27B0} % 摘要框紫色

\renewcommand{\maketitle}{\mymaketitle}

\begin{document}

% 论文内容
\title{ParallelWorld: Test-Time Scaling for Agentic Embodied Reasoning}
\author{Min Chen$^{1,*}$, Shengjun Zhang$^{1,*, \dag}$, Yuxin Li$^{1}$, Zhang Zhang$^{1}$, \\ Xin Fei$^{2}$, Chong Xia$^{1}$, Yueqi Duan$^{1}$}
\affiliation[]{$^{1}$Tsinghua University, $^{2}$National University of Singapore}

\begingroup
\renewcommand{\thefootnote}{}% 清除编号
\footnotetext{$*$ Equal contribution, $\dag$ Project Leader}% 输出无编号的脚注文本
\endgroup

\abstract{
Embodied AI has emerged as a key paradigm for autonomous systems operating in complex physical environments. While end-to-end models have established strong local capabilities for perception and control, scaling these capabilities to extended, multi-stage tasks remains difficult. Agentic embodied systems address this challenge by introducing high-level planners that organize perception, reasoning, and executable skills over longer interaction horizons. However, existing agentic embodied frameworks typically generate actions for direct execution without explicitly simulating how alternative action sequences shape future observations and task progress, leading agents to commit to suboptimal interactions before considering their long-term consequences. To address this limitation, we propose \textbf{ParallelWorld}, an adaptive-horizon test-time scaling framework that grounds agentic embodied reasoning in the prospective simulation of action outcomes. Specifically, we design an agent system to simulate and evaluate alternative trajectories in parallel before committing to an action. A verifier assesses intermediate state transitions with respect to the task objective, dynamically pruning unpromising branches and guiding subsequent expansion. Based on the evaluated long-horizon outcomes, the agent selects a trajectory, executes only its first action, and replans as new observations become available. Experiments on ESI-Bench and LIBERO-Pro demonstrate improvements in active spatial reasoning and language-conditioned manipulation, respectively, highlighting the effectiveness of prospective planning for both understanding and manipulation tasks.
}

\checkdata[
\raisebox{-0.2em}{\includegraphics[width=0.025\linewidth]{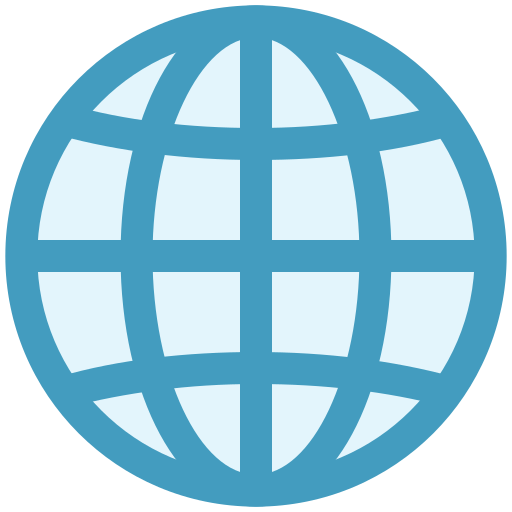}}~~Project Page]{\href{https://agenticei.github.io/ParallelWorld-page/}{\texttt{https://agenticei.github.io/ParallelWorld-page/}}
\\[-1.5ex]}

\checkdata[
\raisebox{-0.2em}{\includegraphics[width=0.025\linewidth]{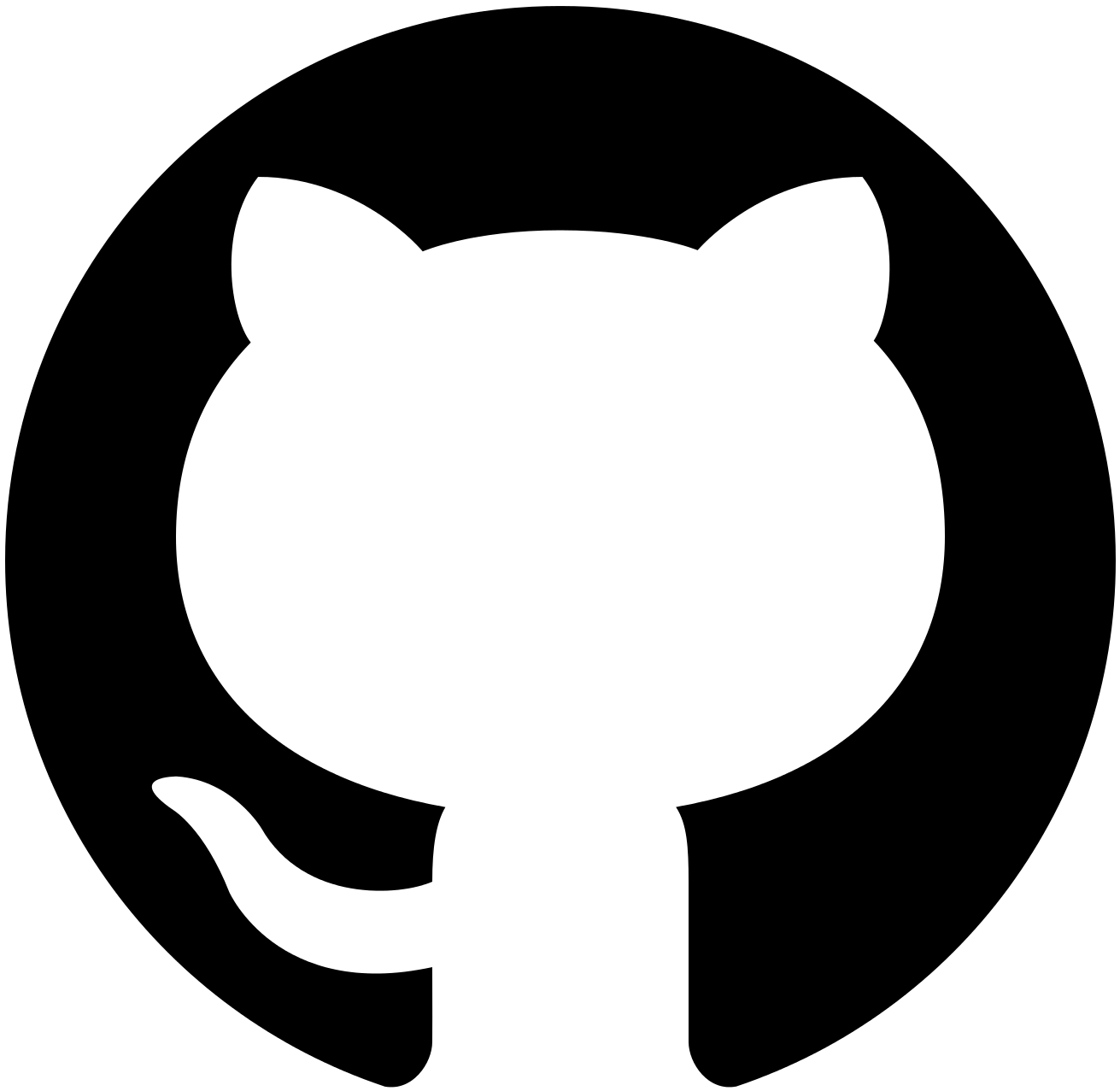}}~~GitHub Repo]{\href{https://github.com/AgenticEI/ParallelWorld}{\texttt{https://github.com/AgenticEI/ParallelWorld}}
\\[-1.5ex]}

% 调用标题+摘要（无报错）
\maketitle

\vspace{1em}

\section{Introduction}

\begin{figure}[t]
    \centering
    \includegraphics[width=\linewidth]{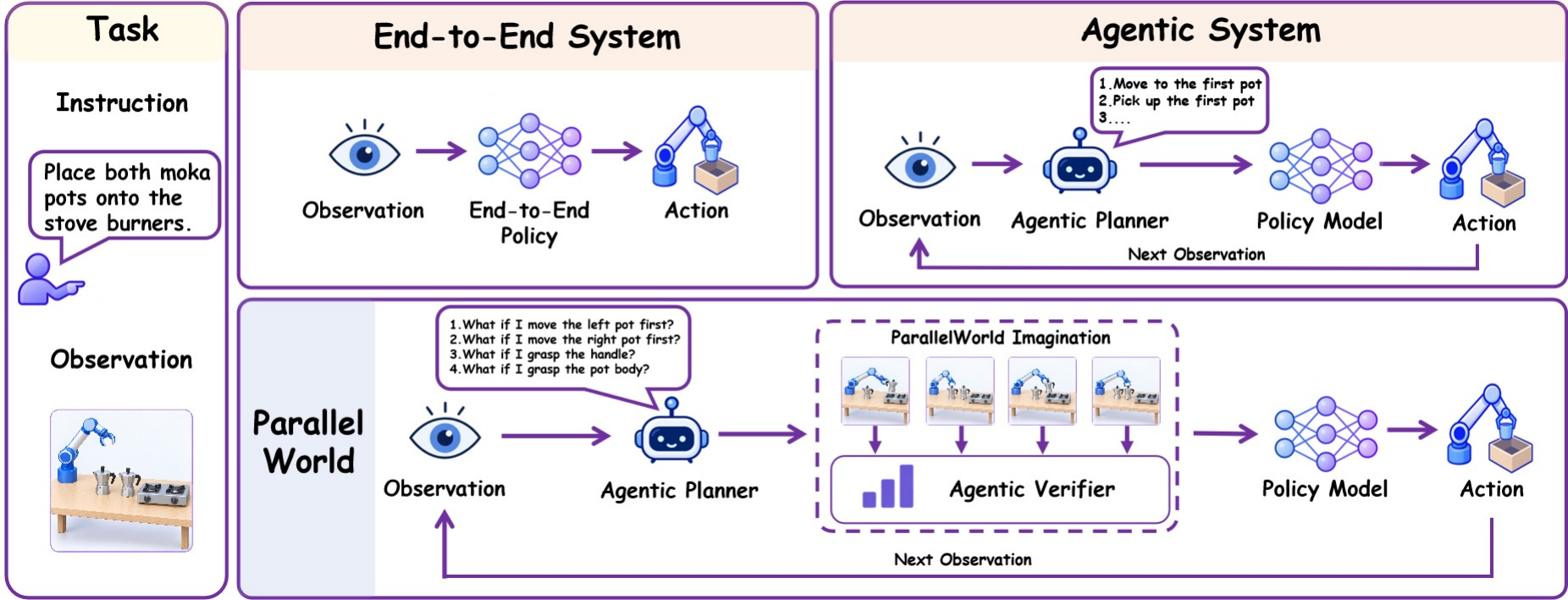}
    \caption{\textbf{Comparison of embodied paradigms.} End-to-end systems directly map observations and instructions to actions, while conventional agentic systems introduce a high-level planner to organize executable skills but still commit to actions without explicit prospective evaluation. ParallelWorld augments agentic planning with multi-step world imagination: the planner proposes alternative interaction trajectories, a verifier evaluates their prospective outcomes, and the selected action is executed before replanning from the next observation.}
    \label{fig:teaser}
\end{figure}

Embodied intelligence enables autonomous systems to integrate perception, reasoning, and action in complex physical environments, supporting tasks such as embodied understanding, navigation, and manipulation~\citep{yang2025embodiedbench,majumdar2024openeqa}. Recent end-to-end approaches have achieved strong performance on embodied understanding from provided observations and on short-horizon interaction tasks, including navigation and manipulation~\citep{chen2024spatialvlm,chen2021history,kim2024openvla,black2024pi_0,intelligence2025pi_}. However, long-horizon tasks require sustained coordination across changing observations, intermediate task states, and successive decisions, which is difficult to achieve through local end-to-end predictions alone. Recent advances in large language and vision-language models have made such coordination increasingly tractable, giving rise to agentic embodied systems that decompose complex objectives and orchestrate executable skills over extended interactions~\citep{liang2023code,hong2026esibench,zhang2026harness}.

As shown in Figure ~\ref{fig:teaser}, current agentic embodied systems typically employ a high-level planner to organize task-dependent interactions over perception and control modules. For embodied understanding, the planner selects exploratory actions to acquire task-relevant evidence, whereas in manipulation it composes executable skills or primitives to achieve a specified physical goal~\citep{hong2026esibench,lei2026towards,zhang2026harness,fu2026cap,liang2023code,lu2026aspire}. In both settings, planning is commonly interleaved with immediate execution, such that the next decision is made only after observing the outcome of the current action without explicitly simulating how alternative action sequences may alter future observations and task progress. As a result, agents can commit to locally plausible interactions before their downstream consequences are sufficiently assessed, causing early decisions to restrict later evidence acquisition or place the environment in states that are less favorable for subsequent execution.

To tackle these challenges, we propose \textbf{ParallelWorld}, an adaptive-horizon test-time scaling framework that augments agentic embodied reasoning with explicit evaluation of future action consequences. Specifically, we construct a search tree through imagination, with each branch representing a candidate action sequence. By extending multiple branches in parallel over several interaction steps, the planner can assess an action in terms of the subsequent observations. A verifier guides this process by assessing the evolving branches against the task objective, retaining promising continuations and pruning less useful alternatives. We couple this prospective search with receding-horizon execution: the agent performs only the first action of the selected trajectory and initiates a new search after observing the resulting state. ParallelWorld directs additional test-time computation toward examining alternative courses of interaction before execution.

We conduct extensive experiments on ESI-Bench~\citep{hong2026esibench} and LIBERO-Pro~\citep{zhou2025libero}, covering embodied understanding and manipulation, respectively. These benchmarks examine two complementary aspects of agentic embodied reasoning: planning interactions to acquire task-relevant evidence and planning actions to achieve a physical goal. Experiments demonstrate improvements in both settings, supporting prospective world simulation as a mechanism for strengthening high-level embodied planning. Our main contributions are summarized as follows:
\begin{itemize}
    \item We propose ParallelWorld, an adaptive-horizon test-time scaling framework that enables agentic embodied reasoning through the prospective simulation of alternative trajectories.
    \item We introduce a verifier-guided tree-search mechanism that evaluates action sequences and couples trajectory selection with receding-horizon execution.
    \item We demonstrate improvements on ESI-Bench and LIBERO-Pro, establishing the effectiveness of prospective planning for both understanding and manipulation embodied tasks.
\end{itemize}

\section{Related Work}

\subsection{LLM-Based Agentic Systems}

Early LLM-based agents developed different ways to connect reasoning with external actions. ReAct~\citep{yao2022react} interleaved reasoning and acting during task solving, while Toolformer~\citep{schick2023toolformer} trained a language model to invoke external tools. AutoGen~\citep{wu2023autogen} supported more complex workflows through programmable coordination among agents. Executable code provided another action interface: CodeAct~\citep{wang2024executable} represented actions as Python programs, enabling compositional tool use and refinement based on execution feedback. In software engineering, SWE-agent~\citep{yang2024swe} introduced interfaces for repository navigation, editing, and testing, while OpenHands~\citep{wang2025openhands} integrated code execution, command-line interaction, and web browsing. Meta-Harness~\citep{lee2026meta} further searched over harness code, using evaluations and execution traces of previous candidates to guide subsequent revisions. These works established interfaces for planning, tool use, and action execution. In embodied tasks, agents must also ground their decisions in visual observations and executable actions within an environment.

% \subsection{Active Embodied Understanding}

% Recent multimodal models strengthen spatial understanding through geometry-aware supervision and representations~\citep{chen2024spatialvlm,cheng2024spatialrgpt,wu2026spatialmllm,fan2026vlm3r}, while complementary work investigates explicit visual reasoning~\citep{wu2026interwoven} and evaluates spatial memory and spatiotemporal understanding from video observations~\citep{yang2025thinking,huang2026thinking}. Beyond reasoning over provided observations, active embodied understanding incorporates environment interaction into the acquisition of task-relevant evidence, building on active perception~\citep{bajcsy1988active,aloimonos1988active} and the active-exploration formulation of embodied question answering~\citep{das2018embodied,majumdar2024openeqa}. Recent approaches improve this process through confidence-calibrated exploration~\citep{ren2024explore}, retrospective experience replay and hierarchical frontier selection~\citep{zhang2025reexplore}, and closed-loop visual reasoning through physical interaction~\citep{zhou2025physvlmavr}. Theory of Space examines the construction and revision of spatial beliefs during autonomous exploration~\citep{zhang2026theory}, whereas ESI-Bench evaluates how agents coordinate perception, locomotion, and manipulation to resolve spatial questions~\citep{hong2026esibench}. 
\subsection{Agentic Embodied Systems}

Recent embodied systems increasingly rely on high-level agents to structure perception, reasoning, and physical interaction into goal-directed behaviors. For embodied understanding, earlier studies improve spatial reasoning from given observations through geometry-aware representations and supervision~\citep{chen2024spatialvlm,cheng2024spatialrgpt,wu2026spatialmllm,fan2026vlm3r}, while active approaches acquire additional task-relevant evidence through exploration or physical interaction~\citep{das2018embodied,majumdar2024openeqa,ren2024explore,zhou2026physvlm,hong2026esibench}. In robotic manipulation, agentic systems translate language instructions into executable programs or high-level primitives, as exemplified by Code as Policies~\citep{liang2023code}, CaP-X~\citep{fu2026cap}, ASPIRE~\citep{lu2026aspire}, VLAs-as-Tools~\citep{lei2026towards}, and Harness VLA~\citep{zhang2026harness}. Although these two lines of work differ in their immediate objectives, both rely on high-level agents to organize sequences of interactions according to the task context: understanding-oriented agents act to acquire informative evidence, whereas manipulation-oriented agents act to transform the environment toward a desired goal. This common structure motivates prospective reasoning over alternative action outcomes before execution, which forms the basis of ParallelWorld.

\subsection{Embodied Test-Time Scaling}

Test-time scaling improves model inference by allocating additional computation at deployment through candidate generation, search, and verification. Early language-model approaches generate multiple candidate solutions and rank them with learned verifiers~\citep{cobbe2021training}, while subsequent work extends this paradigm through structured search~\citep{yao2023tree} and adaptive allocation of inference-time computation~\citep{snell2025scaling}.

Recent studies have extended test-time scaling to embodied intelligence, where additional computation is used to improve action selection before execution. RoboMonkey~\citep{kwok2025robomonkey} samples multiple VLA actions and selects among them using a vision-language verifier, while RoVer~\citep{dai2025rover} introduces a robot process reward model to score and iteratively refine candidate actions. TACO~\citep{yang2025steering} improves inference stability by selecting reliable action chunks from multiple VLA samples using a lightweight pseudo-count verifier, whereas E-TTS~\citep{ye2026tts} jointly scales reasoning and action generation through history-aware verification and iterative refinement. Beyond direct action verification, $\omega$-EVA~\citep{sun2026omegaeva} incorporates proposal-conditioned future consequences into action refinement, and Gated GeoBoN~\citep{zhao2026geobon} selectively allocates additional rollout computation and ranks candidate rollouts according to action--future and geometric consistency. ParallelWorld advances this line of research by introducing prospective reasoning over alternative high-level embodied interactions before execution.

\section{Methods}

\begin{figure}[t]
    \centering
    \includegraphics[width=\linewidth]{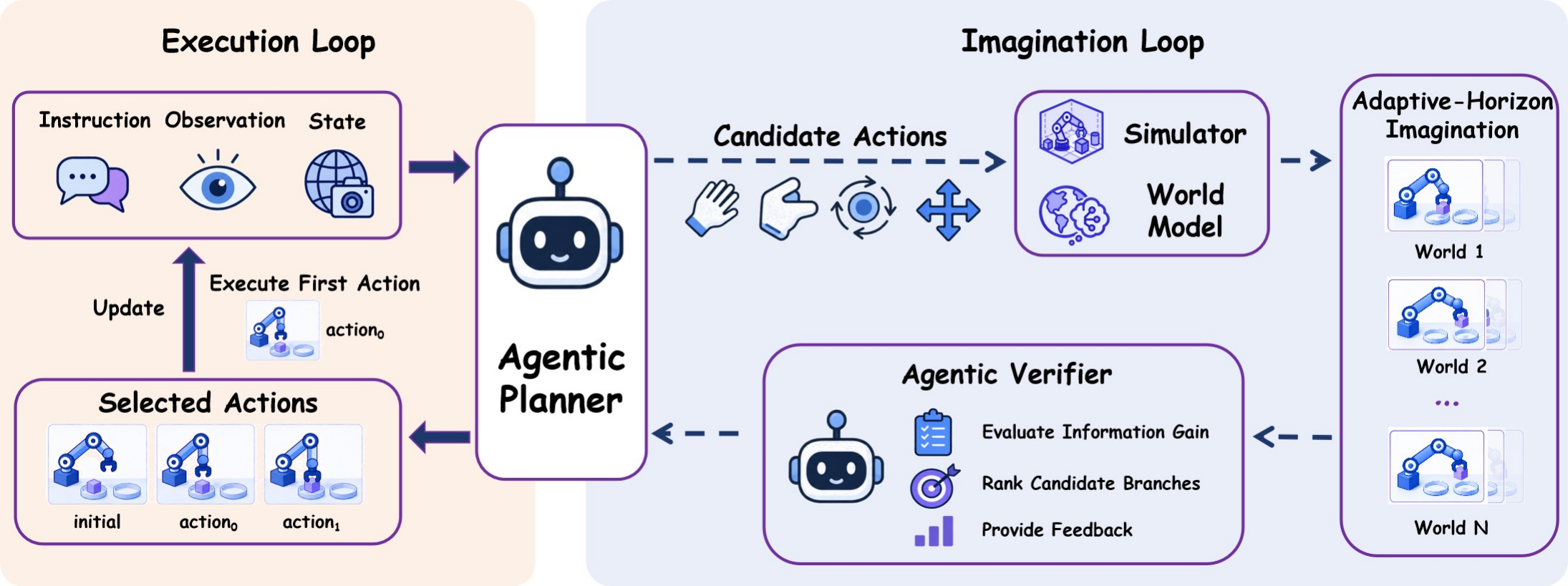}
    \caption{
        \textbf{Overview of ParallelWorld.}
        Given a task instruction and the current observation, an agentic
        planner explores alternative actions through simulator-based or
        world-model-based imagination. A verifier selects promising
        branches, while the planner determines whether to continue
        imagination or commit to execution. Only the first action of the
        selected trajectory is executed, followed by replanning from the
        resulting observation. Dashed and solid arrows denote imagination
        and environment interaction, respectively.
    }
    \label{fig:pipeline}
\end{figure}

\subsection{Problem Formulation}

We consider agentic embodied reasoning as high-level planning in a partially observed environment. Given a language input $\ell$ and the current observation $o_t$, the planner selects executable high-level actions. An interaction episode yields a sequence:
\begin{equation}
    \tau =
    \left(
        a_0,a_1,\ldots,a_{T-1}
    \right),
    \qquad
    a_t=(p_t,\theta_t)\in\mathcal{A},
\end{equation}
where $p_t$ denotes an atomic action primitive and $\theta_t$ its associated arguments. Each primitive defines a bounded unit of interaction with a well-defined execution scope and termination condition. The planner therefore operates over semantic actions rather than directly predicting low-level trajectories.

Executing a primitive induces a transition in the physical environment,
\begin{equation}
    s_{t+1}
    =
    \mathcal{T}(s_t,a_t),
    \qquad
    o_{t+1}
    =
    \mathcal{O}(s_{t+1}),
\end{equation}
where $s_t$ denotes the underlying environment state and $o_t$ the
resulting observation. The primitive space may combine perceptual and
physical interactions. For example, camera motion and object interaction
form the action space for embodied understanding, while robotic
manipulation is expressed through a fixed set of analytic and VLA-backed
primitives~\citep{zhang2026harness}.

Within this shared formulation, the two task settings differ primarily
in their objectives. For understanding, the language input is a question
$q$, and the agent interacts with the environment to acquire evidence
$\mathcal{E}_{1:T}$ for predicting an answer:
\begin{equation}
    \hat{y}
    =
    A
    \left(
        q,\mathcal{E}_{1:T}
    \right).
\end{equation}
For manipulation, the language input is an instruction $g$. Repeated
application of the single-step transition $\mathcal{T}$ should bring the
environment to a terminal state satisfying the task goal:
\begin{equation}
    s_T\in\mathcal{G}(g),
\end{equation}
where $\mathcal{G}(g)$ denotes the state consistent with $g$. The overall framework is illustrated in Figure ~\ref{fig:pipeline}.

\subsection{World Imagination}

Before committing to an action, ParallelWorld constructs multiple
prospective worlds to evaluate alternative courses of interaction.
Starting from the current environment observation $o_t$, the imagination
process proceeds recursively over prospective steps indexed by $h$,
which are distinguished from actual environment interactions indexed
by $t$.

We represent a prospective world as:
\begin{equation}
    b_h=(o_h,\xi_h),
\end{equation}
where $o_h$ is its prospective observation and $\xi_h$ is an optional restorable simulator state. We write $o(b_h)=o_h$ and $\xi(b_h)=\xi_h$ for its two components. For simulator-based imagination, $\xi_h$ retains the state needed to resume that branch; for learned-model imagination, $\xi_h=\varnothing$ because candidate outcomes are generated without restoring a simulator state. The root branch is:
\begin{equation}
    \mathcal{B}_0=\{b_0\},
    \qquad
    b_0=(o_t,\xi_t),
\end{equation}
where $\xi_t$ is the current simulator snapshot when simulator-based imagination is used.

At imagination step $h$, the planner proposes a set of candidate
primitive actions conditioned on the language objective and the
current prospective worlds:
\begin{equation}
    \mathcal{C}_h
    =
    \Pi_{\mathrm{cand}}
    \left(
        \ell,
        \mathcal{B}_{h-1}
    \right),
    \qquad
    K_h=|\mathcal{C}_h|,
\end{equation}
where $\mathcal{B}_{h-1}$ denotes the prospective worlds retained from
the previous step, and $K_h$ is the search width determined by the
planner. It may consider several alternative actions when further
comparison is necessary, while proposing fewer candidates when the
next interaction is sufficiently clear. Each candidate
\begin{equation}
    c_h^{(i)}
    =
    \left(
        b_{h-1}^{p(i)},
        a_h^{(i)}
    \right),
    \qquad i=1,\ldots,K_h,
\end{equation}
associates an atomic primitive $a_h^{(i)}$ with one retained parent
world $b_{h-1}^{p(i)}$. Here $p(i)$ identifies the parent of candidate
$i$.

Each candidate is mapped to a prospective successor through a
world-imagination operator:
\begin{equation}
    \tilde{b}_h^{(i)}
    =
    \mathcal{W}
    \left(
        b_{h-1}^{p(i)},
        a_h^{(i)};
        \ell
    \right).
\end{equation}
The implementation of $\mathcal{W}$ depends on the task setting.
For understanding, the task often requires acquiring evidence that
is not observable from the current view. Predicting such content
directly from the available observation would require a learned
world model to infer the very information being sought. We therefore
obtain prospective outcomes from the simulator:
\begin{equation}
    \xi_h^{(i)}
    =
    \mathcal{T}
    \left(
        \xi\bigl(b_{h-1}^{p(i)}\bigr),
        a_h^{(i)}
    \right),
    \qquad
    \tilde{b}_h^{(i)}
    =
    \left(
        \mathcal{O}\bigl(\xi_h^{(i)}\bigr),
        \xi_h^{(i)}
    \right).
\end{equation}
Before evaluating a candidate, its parent simulator state is restored.
This allows different actions to be evaluated from the same parent
world. The successor retains both its resulting observation and the
state needed for further expansion.

For manipulation, we instead use an image generation model to
predict the immediate visual outcome of each primitive. Given a
proposed primitive $a_h^{(i)}$, we first convert it into a
natural-language action prompt:
\begin{equation}
    \sigma_h^{(i)}
    =
    \mathcal{F}
    \left(
        \ell,
        o\bigl(b_{h-1}^{p(i)}\bigr),
        a_h^{(i)}
    \right).
\end{equation}
The image model then generates the corresponding post-action
observation:
\begin{equation}
    \tilde{b}_h^{(i)}
    =
    \left(
        G_{\psi}
        \left(
            o\bigl(b_{h-1}^{p(i)}\bigr),
            \sigma_h^{(i)}
        \right),
        \varnothing
    \right).
\end{equation}
The generated observation serves as the starting point for
subsequent prospective expansion.

The resulting candidate worlds
\begin{equation}
    \widetilde{\mathcal{B}}_h
    =
    \left\{
        \tilde{b}_h^{(1)},
        \ldots,
        \tilde{b}_h^{(K_h)}
    \right\}
\end{equation}
are jointly evaluated by a verifier:
\begin{equation}
    \left(
        \mathcal{B}_h,
        \rho_h^{*},
        f_h
    \right)
    =
    V_{\phi}
    \left(
        \ell,
        \widetilde{\mathcal{B}}_h
    \right),
    \qquad
    M_h=|\mathcal{B}_h|,
\end{equation}
where $\mathcal{B}_h\subseteq\widetilde{\mathcal{B}}_h$ denotes the
prospective worlds retained for further expansion, $\rho_h^{*}$
is the verifier's currently preferred root-to-leaf trajectory,
and $f_h$ denotes its comparative feedback. The verifier evaluates
the observations associated with the candidates; simulator states
are retained internally for branch restoration rather than presented
as visual evidence. Parent links are preserved so that the preferred
trajectory can be traced from its selected leaf to the root. The
retention width $M_h$ is determined by the verifier after observing
the prospective outcomes.

Finally, the planner determines whether another imagination step is
necessary:
\begin{equation}
    z_h
    =
    \Pi_{\mathrm{ctrl}}
    \left(
        \ell,
        \mathcal{B}_h,
        \rho_h^{*},
        f_h
    \right),
    \qquad
    z_h\in
    \{\mathrm{continue},\mathrm{stop}\}.
\end{equation}
If $z_h=\mathrm{continue}$, the retained worlds become the parents
of the next prospective step and the planner generates a new
candidate set $\mathcal{C}_{h+1}$; otherwise, imagination
terminates and the selected trajectory $\rho_h^{*}$ is passed to
the interaction stage.

\subsection{Foresight-Guided Interaction}

When the planner terminates world imagination, the verifier-selected
trajectory determines the next environment interaction. Let $H$
denote the imagination depth at termination, and let
\begin{equation}
    \rho_H^{*}
    =
    \left(
        a_{t,1}^{*},
        a_{t,2}^{*},
        \ldots,
        a_{t,H}^{*}
    \right)
\end{equation}
denote the preferred trajectory returned by the verifier, where
$t$ indexes the current environment interaction step and
$a_{t,j}^{*}$ is the $j$-th imagined primitive along the trajectory.

Rather than executing the entire imagined trajectory, ParallelWorld
commits only its first action:
\begin{equation}
    \hat{a}_t=a_{t,1}^{*}.
\end{equation}
The selected primitive is executed in the environment, yielding
the next state and observation:
\begin{equation}
    s_{t+1}
    =
    \mathcal{T}(s_t,\hat{a}_t),
    \qquad
    o_{t+1}
    =
    \mathcal{O}(s_{t+1}).
\end{equation}

By executing only the first action, ParallelWorld avoids committing
to an entire imagined trajectory whose later states may become
increasingly inaccurate, while preserving the benefit of multi-step
lookahead for current action. The newly obtained observation then initiates the next planning
cycle.

\section{Experiments}

\begin{table*}[t!]
\centering
\caption{
\textbf{Quantitative results on ESI-Bench~\citep{hong2026esibench}.} 
We report accuracy for Passive, Active Exploration, and ParallelWorld. All active methods use the same answer-model. Row colors follow the ESI-Bench taxonomy: yellow denotes object representation, purple denotes layout and geometry, blue denotes number representation, and green denotes agents and goal-directed actions.}
\label{tab:understanding}

\begingroup
\scriptsize
\setlength{\tabcolsep}{6pt}
\renewcommand{\arraystretch}{1.25}
\newcommand{\resdash}{\textcolor{black!45}{--}}

\begin{tabularx}{\textwidth}{
    >{\raggedright\arraybackslash}p{0.24\textwidth}
    >{\raggedright\arraybackslash}X
    >{\centering\arraybackslash}p{0.14\textwidth}
    >{\centering\arraybackslash}p{0.14\textwidth}
    >{\centering\arraybackslash}p{0.14\textwidth}
}
\toprule
\multirow{2}{*}{\footnotesize Category} &
\multirow{2}{*}{\footnotesize Subcategory} &
\multicolumn{3}{c}{\footnotesize Agent Evaluation} \\
\cmidrule(lr){3-5}
& &
\footnotesize Passive &
\footnotesize Active &
\footnotesize ParallelWorld \\
\midrule

% =========================================================
% Perceptual Grounding (3行)
% =========================================================
\rowcolor{pwYellow}
& \textit{View Hallucination}
& 53.32\% & 66.82\% & \textbf{71.40\%}  \\
\rowcolor{pwYellow}
& \textit{Partial Occlusion}
& 36.84\% & 57.89\% & \textbf{77.89\%}  \\
\rowcolor{pwYellow}
\multirow{-3}{*}{\footnotesize{Perceptual Grounding}}
& \textit{Material Transparency}
& 69.72\% & 71.10\% & \textbf{73.85\%}  \\
\midrule

% =========================================================
% Physical Structure (3行)
% =========================================================
\rowcolor{pwYellow}
& \textit{Rigid Containment}
& 55.00\% & \textbf{60.00\%} & 50.00\%  \\
\rowcolor{pwYellow}
\multirow{-2}{*}{\footnotesize{Physical Structure}}
& \textit{Deformable Objects}
& 37.76\% & 40.82\% & \textbf{42.86\%}  \\
\midrule

% =========================================================
% Physical Dynamics (2行)
% =========================================================
\rowcolor{pwYellow}
& \textit{Inclined Plane}
& 59.02\% & 83.61\% & \textbf{85.25\%}  \\
\rowcolor{pwYellow}
\multirow{-2}{*}{\footnotesize{Physical Dynamics}}
& \textit{Stacking \& Stability}
& 1.12\% & 45.98\% & \textbf{53.95\%}  \\
\midrule

% =========================================================
% Specular Reflection (3行)
% =========================================================
\rowcolor{pwYellow}
& \textit{Reflection Authoring}
& 56.57\% & 51.52\% & \textbf{60.61\%}  \\
\rowcolor{pwYellow}
& \textit{Spatial Relations}
& 38.94\% & 36.28\% & \textbf{51.33\%}  \\
\rowcolor{pwYellow}
\multirow{-3}{*}{\footnotesize{Specular Reflection}}
& \textit{Correspondence}
& 42.05\% & 47.73\% & \textbf{48.86\%}  \\
\midrule

% =========================================================
% Spatial Relations (3行)
% =========================================================
\rowcolor{pwPurple}
& \textit{Linear Alignment}
& 52.13\% & 68.09\% & \textbf{70.21\%} \\
\rowcolor{pwPurple}
& \textit{Geometric Configuration}
& 16.55\% & \textbf{17.96\%} & 16.90\% \\
\rowcolor{pwPurple}
\multirow{-3}{*}{\footnotesize{Spatial Relations}}
& \textit{Physical Contact}
& 66.39\% & 68.07\% & \textbf{73.95\%} \\
\midrule

% =========================================================
% Metric Comparison (2行)
% =========================================================
\rowcolor{pwPurple}
& \textit{Dimensional Size}
& 39.52\% & 44.91\% & \textbf{49.10\%} \\
\rowcolor{pwPurple}
\multirow{-2}{*}{\footnotesize{Metric Comparison}}
& \textit{Spatial Distance}
& 61.18\% & 58.55\% & \textbf{67.11\%} \\
\midrule

% =========================================================
% Cognitive Mapping (4行)
% =========================================================
\rowcolor{pwPurple}
& \textit{Connectivity}
& \textbf{66.67\%} & 55.00\% & 60.00\% \\
\rowcolor{pwPurple}
& \textit{Traversable Passage}
& 68.33\% & 65.00\% & \textbf{75.00\%} \\
\rowcolor{pwPurple}
& \textit{Regional Boundary}
& 45.00\% & 57.50\% & \textbf{63.75\%} \\
\rowcolor{pwPurple}
\multirow{-4}{*}{\footnotesize{Cognitive Mapping}}
& \textit{Long-Term Navigation}
& 16.67\% & 16.67\% & \textbf{18.33\%} \\
\midrule

% =========================================================
% Enumerative Perception (6行)
% =========================================================
\rowcolor{pwBlue}
& \textit{Counting with Occlusion}
& 23.33\% & 26.67\% & \textbf{30.00\%} \\
\rowcolor{pwBlue}
& \textit{Spatial Segmentation}
& 20.00\% & 20.00\% & \textbf{23.33\%} \\
\rowcolor{pwBlue}
& \textit{Merged Observation}
& 0.00\% & 1.67\% & \textbf{6.67\%} \\
\rowcolor{pwBlue}
& \textit{Category Ambiguity}
& 0.00\% & 8.33\% & \textbf{13.33\%} \\
\rowcolor{pwBlue}
& \textit{Structural Enclosure}
& 2.50\% & 5.00\% & \textbf{7.50\%} \\
\rowcolor{pwBlue}
\multirow{-6}{*}{\footnotesize{Enumerative Perception}}
& \textit{Illumination Variability}
& 0.00\% & 6.00\% & \textbf{10.00\%} \\
\midrule

% =========================================================
% Temporal Understanding (2行)
% =========================================================
\rowcolor{pwGreen}
& \textit{Unobserved Change}
& 38.62\% & 70.95\% & \textbf{89.86\%} \\
\rowcolor{pwGreen}
\multirow{-2}{*}{\footnotesize{Temporal Understanding}}
& \textit{Agent Observation}
& 34.59\% & 48.87\% & \textbf{62.41\%} \\
\midrule

% =========================================================
% Action Sequencing (1行，无需multirow)
% =========================================================
\rowcolor{pwGreen}
\footnotesize{Action Sequencing}
& \textit{Action Order Inference}
& 37.66\% & 53.25\% & \textbf{61.04\%} \\

\bottomrule
\end{tabularx}
\endgroup
\end{table*}

\label{sec:experiments}

We evaluate ParallelWorld on embodied understanding and manipulation. We first evaluate the effectiveness of prospective world imagination for embodied understanding in Section~\ref{sec:exp_understanding}. In Section~\ref{sec:exp_manipulation}, we further assess ParallelWorld on robotic manipulation, examining whether imagined action outcomes improve high-level planning and task execution. In Section~\ref{sec:exp_ablation}, we conduct an ablation study on the search width $K_h$ in ParallelWorld.

\subsection{Embodied Understanding}
\label{sec:exp_understanding}

\subsubsection{Experimental Setup}
\paragraph{Benchmark.} We conduct experiments on ESI-Bench~\citep{hong2026esibench}, a comprehensive benchmark for embodied spatial intelligence. The benchmark contains 10 task categories and 29 subcategories. Due to an environment issue that prevents the execution of pouring actions, we exclude the Liquid Volume subcategory and evaluate the remaining 28 subcategories. 

\paragraph{Baselines.} We selected two paradigms from ESI-Bench~\citep{hong2026esibench} as baselines, including Passive Single-View, which predicts the answer from a single observation at the initial pose, and Active Exploration, which sequentially selects camera or physical actions to gather task-relevant evidence before answering.

\paragraph{Implementation Details.} We build our implementation upon the active exploration framework provided by ESI-Bench~\citep{hong2026esibench}. The planner iteratively decides whether to continue exploration based on the confidence of the current prediction. Specifically, exploration terminates when the answer confidence exceeds 0.8 or when the maximum exploration depth of 15 is reached.

\subsubsection{Main Results}
As shown in Table~\ref{tab:understanding}, ParallelWorld consistently improves over Active Exploration on the majority of ESI-Bench subcategories, demonstrating the effectiveness of prospective world imagination for embodied understanding. The gains are particularly evident on tasks that require additional interaction to reveal task-relevant evidence or resolve uncertainty, where the accuracy on Unobserved Change increases from 70.95\% to 89.86\% and Partial Occlusion improves from 57.89\% to 77.89\%. Compared with sequential action selection, ParallelWorld benefits from evaluating multiple possible future trajectories before execution, leading to more effective decisions across diverse spatial reasoning tasks. We also observe several subcategories where the improvement is limited or performance slightly decreases compared with the baselines. These cases are further analyzed in Section~\ref{sec:discussions}.

% Table~\ref{tab:main_results} summarizes the quantitative results across the 28 evaluated ESI-Bench subcategories. ParallelWorld consistently outperforms the conventional Active Exploration baseline under the same answer-model backbone. The improvements are particularly pronounced for Temporal Understanding, where the accuracy on Unobserved Change increases from 70.95\% to 89.86\%, and for Metric Comparison, where Spatial Distance improves from 58.55\% to 67.11\%. Although Passive Single-View performs best on Connectivity due to the task setting, ParallelWorld still improves over Active Exploration from 55.00\% to 60.00\%. Overall, these results demonstrate the effectiveness of prospective simulation for active embodied reasoning.

\begin{table*}[t!]
\centering
\caption{
\textbf{Quantitative results on LIBERO-Pro~\citep{zhou2025libero}.}
We report success rates (\%) under instruction-redirection (T) and position-swap (S) perturbations across the \textsc{Spatial}, \textsc{Object}, \textsc{Goal}, and \textsc{LIBERO-10} suites. Each setting contains 100 trials (10 tasks $\times$ 10 seeds).}
\label{tab:manipulation}

\begin{tabular}{lcccccccc}
\toprule
\multirow{2}{*}{Method}
& \multicolumn{2}{c}{Spatial}
& \multicolumn{2}{c}{Object}
& \multicolumn{2}{c}{Goal}
& \multicolumn{2}{c}{LIBERO-10} \\
\cmidrule(lr){2-3}
\cmidrule(lr){4-5}
\cmidrule(lr){6-7}
\cmidrule(lr){8-9}
& T & S & T & S & T & S & T & S \\
\midrule

OpenVLA~\citep{kim2024openvla}
& 0.0 & 0.0
& 0.0 & 0.0
& 0.0 & 0.0
& 0.0 & 0.0 \\

$\pi_0$~\citep{black2024pi_0}
& 0.0 & 0.0
& 0.0 & 2.0
& 0.0 & 0.0
& 0.0 & 0.0 \\

$\pi_{0.5}$~\citep{intelligence2025pi_}
& 1.0 & 20.0
& 1.0 & 17.0
& 2.0 & 38.0
& 1.0 & 8.0 \\

MolmoAct~\citep{lee2025molmoact}
& 0.0 & 0.0
& 0.0 & 6.0
& 0.0 & 0.0
& 6.0 & 0.0 \\

NORA~\citep{hung2025nora}
& 0.0 & 0.0
& 0.0 & 0.0
& 0.0 & 0.0
& 0.0 & 0.0 \\

X-VLA~\citep{zheng2026x}
& 0.0 & 0.0
& 8.0 & 2.0
& 9.0 & 1.0
& 10.0 & 0.0 \\

AtomVLA~\citep{sun2026atomvla}
& 1.0 & 16.0
& 0.0 & 10.0
& 11.0 & 2.0
& 9.0 & 1.0 \\

\midrule

HarnessVLA~\citep{zhang2026harness}
& 79.0 & 56.0
& 79.0 & 85.0
& 62.0 & 57.0
& 42.0 & 29.0 \\

\textbf{ParallelWorld (Ours)}
& \textbf{88.0} & \textbf{83.0}
& \textbf{94.0} & \textbf{88.0}
& \textbf{65.0} & \textbf{66.0}
& \textbf{50.0} & \textbf{30.0} \\

\bottomrule
\end{tabular}
\end{table*}
\subsection{Robotic Manipulation}
\label{sec:exp_manipulation}

\subsubsection{Experimental Setup}
\paragraph{Benchmark.} We evaluate manipulation performance on LIBERO-Pro~\citep{zhou2025libero}, an extension of LIBERO designed to assess robotic generalization under perturbations rather than the standard in-distribution setting. We consider the \textsc{Spatial}, \textsc{Object}, \textsc{Goal} and \textsc{LIBERO-10} suites under instruction- redirection (\textbf{T}) and position swap (\textbf{S}) perturbation. Each evaluation cell contains 10 tasks evaluated over 10 seeds, resulting in 100 episodes per setting and 800 episodes in total.

\paragraph{Baselines.} We compare ParallelWorld with representative end-to-end VLA models and agentic manipulation methods. The VLA baselines include OpenVLA~\citep{kim2024openvla}, $\pi_0$~\citep{black2024pi_0}, $\pi_{0.5}$~\citep{intelligence2025pi_}, MolmoAct~\citep{lee2025molmoact}, NORA~\citep{hung2025nora}, X-VLA~\citep{zheng2026x}, and AtomVLA~\citep{sun2026atomvla}. We use HarnessVLA~\citep{zhang2026harness} as the direct agentic baseline and reproduce it with Codex as the agentic planner, whereas the results of the remaining frozen VLA baselines are taken from the original HarnessVLA paper.

\paragraph{Implementation Details.} Our manipulation implementation is built upon the HarnessVLA~\citep{zhang2026harness} framework. For visual imagination, we use GPT-image-2.5 as the image generation model with low image quality settings to predict the prospective outcomes of candidate primitives.

\subsubsection{Main Results}
As shown in Table~\ref{tab:manipulation}, ParallelWorld surpasses HarnessVLA~\citep{zhang2026harness} across all LIBERO-Pro evaluation settings, validating the effectiveness of prospective world imagination for robotic manipulation. The largest gains are observed on the Spatial and Object suites, where ParallelWorld substantially improves the success rates over the agentic baseline, including 56.0\% to 83.0\% on Spatial position-swap and 79.0\% to 94.0\% on Object instruction-redirection. Improvements on the Goal and LIBERO-10 suites are comparatively more moderate, although ParallelWorld remains consistently stronger across all corresponding settings. We further analyze the variation in performance gains across different task suites in Section~\ref{sec:discussions}.

\begin{figure}[t]
    \centering
    \includegraphics[width=\linewidth]{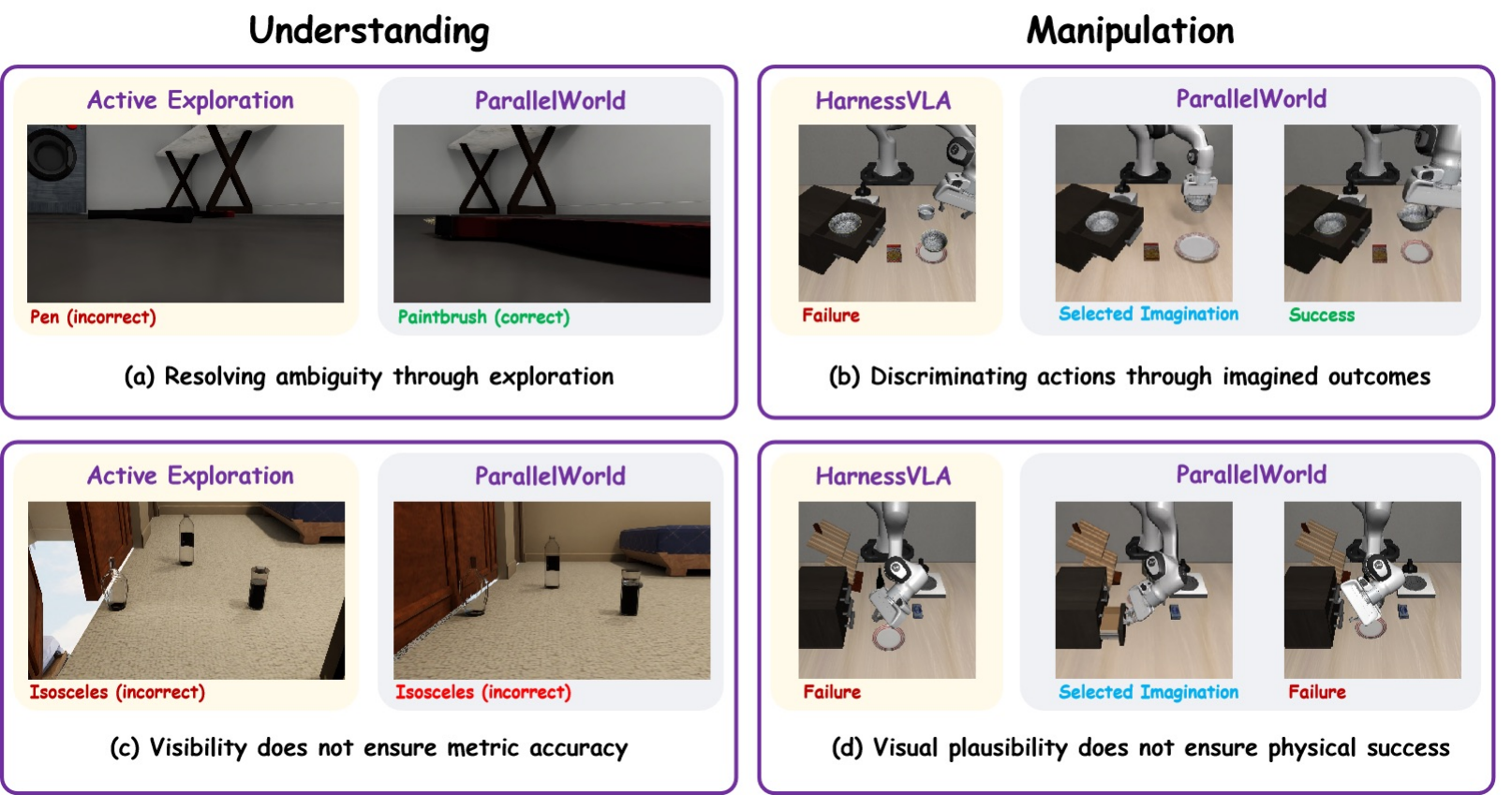}
    \caption{\textbf{Qualitative analysis of ParallelWorld.}
(a) \textbf{Prospective exploration.} Active Exploration selects an uninformative view and fails, while ParallelWorld discovers the discriminative observation through future interaction. 
(b) \textbf{Prospective manipulation.} HarnessVLA~\citep{zhang2026harness} fails with direct execution, whereas ParallelWorld selects an imagined outcome that guides the robot toward a better grasping configuration and achieves success. 
(c) \textbf{Limitation in geometric reasoning.} ParallelWorld obtains sufficient observations, but the model still fails to infer precise metric relationships between objects. 
(d) \textbf{Limitation in physical prediction.} The imagined outcome appears plausible, but visual imagination cannot fully capture contact dynamics required for successful physical execution.}
    \label{fig:discussion}
\end{figure}

\subsection{Discussions}
\label{sec:discussions}
\paragraph{When does ParallelWorld help?} ParallelWorld is most effective when task success depends on discriminating among alternative action outcomes and when the value of an action emerges through its downstream consequences. This is particularly evident in spatial and object-centric manipulation, where candidate actions differ in target identity, relative configuration, or placement geometry, as well as in embodied understanding tasks that require additional interaction to acquire otherwise unavailable evidence. By explicitly comparing prospective worlds before execution, ParallelWorld enables the planner to select interactions with higher expected information gain or task progress rather than relying on the most plausible immediate action. As illustrated in Figure~\ref{fig:discussion}, ParallelWorld can identify informative viewpoints in understanding tasks and select more favorable manipulation outcomes before execution, reducing errors caused by incorrect exploration or action selection.

\paragraph{Limitations.} The effectiveness of ParallelWorld is limited when the decisive factors are not fully reflected in visual outcomes. As shown in Figure~\ref{fig:discussion}, the method may still fail when the underlying challenge requires precise metric reasoning or accurate prediction of physical interactions. For example, although sufficient visual evidence is acquired, current agentic models may struggle with geometric relationships, while image-based imagination may not faithfully capture contact dynamics and execution uncertainty. These limitations are also reflected in Goal and LIBERO-10 tasks, where articulated states, contact-dependent transitions, and long-horizon error accumulation play a larger role. Overall, the effectiveness of prospective imagination depends on whether task-relevant consequences can be reliably represented in the imagined observations.

\subsection{Ablation Studies}
\label{sec:exp_ablation}

We investigate the effect of adaptive search width by comparing ParallelWorld with fixed search width strategies on three Spatial-S tasks with 30 episodes in total. As shown in Table~\ref{tab:ablation}, direct execution achieves only $23.3\%$ success rate, demonstrating the difficulty of solving these tasks without prospective evaluation. Fixed search width strategies provide limited improvements, where $K_h=2$ achieves 10.0\% success rate and $K_h=4$ improves to $36.7\%$ with higher computational cost.

The inferior performance of fixed search widths stems from their inability to adapt the number of candidate worlds to the uncertainty of the current decision. A small fixed width may discard informative candidates when multiple alternatives need to be explored, while a large fixed width introduces unnecessary candidates in relatively deterministic states, increasing both computation and the difficulty of trajectory selection. In contrast, Adaptive Search Width dynamically determines $K_h$ according to the current task state and planning context, allocating more search capacity when alternative futures are ambiguous and reducing redundant exploration otherwise. It achieves a substantially higher success rate of $73.3\%$ while maintaining lower runtime than the fixed $K_h=4$ strategy, demonstrating the importance of adaptive search width for effective test-time planning.

\begin{table}[t]
\centering
\caption{
\textbf{Ablation of adaptive search width on Spatial-S.}
Results are averaged over three tasks with 10 seeds each.
Imagination Time and Total Time represent the average time cost of imagination and the entire episode, respectively.
}
\label{tab:ablation}
\small
\setlength{\tabcolsep}{8pt}
\begin{tabular}{lccc}
\toprule
Method
& Success Rate (\%) $\uparrow$
& Total Time (s) $\downarrow$ \\
\midrule
Direct Execution~\citep{zhang2026harness}
& 23.3
& 417.4 \\

Fixed Search Width ($K_h=2$)
& 10.0
& 2930.0 \\

Fixed Search Width ($K_h=4$)
& 36.7
& 4889.5 \\

\midrule
Adaptive Search Width (Ours)
& \textbf{73.3}
& \textbf{2402.2} \\
\bottomrule
\end{tabular}
\end{table}

\section{Conclusion}

We presented ParallelWorld, an adaptive-horizon test-time scaling framework for agentic embodied reasoning. Rather than directly committing to locally plausible actions, ParallelWorld explicitly explores alternative prospective worlds before execution, allowing future observations and task consequences to inform high-level decisions. We instantiate ParallelWorld for both embodied understanding and robotic manipulation, using simulator-based transitions and image-based outcome prediction, respectively. Experiments demonstrate consistent improvements over existing active and agentic baselines, while ablation studies further validate the effectiveness of adaptive search width. These results establish prospective world imagination as an effective mechanism for extending agentic embodied systems from reactive action selection toward foresight-driven decision making.

\bibliographystyle{plain}
\bibliography{main}

@article{kwok2025robomonkey,
  title={Robomonkey: Scaling test-time sampling and verification for vision-language-action models},
  author={Kwok, Jacky and Agia, Christopher and Sinha, Rohan and Foutter, Matt and Li, Shulu and Stoica, Ion and Mirhoseini, Azalia and Pavone, Marco},
  journal={arXiv preprint arXiv:2506.17811},
  year={2025}
}

@article{dai2025rover,
  title={Rover: Robot reward model as test-time verifier for vision-language-action model},
  author={Dai, Mingtong and Liu, Lingbo and Bai, Yongjie and Liu, Yang and Wang, Zhouxia and Su, Rui and Chen, Chunjie and Lin, Liang and Wu, Xinyu},
  journal={arXiv preprint arXiv:2510.10975},
  year={2025}
}

@article{zhao2026geobon,
  title={Test-Time Scaling for World Action Models via Zero-Shot Geometric Evaluation},
  author={Zhao, Zesen and Cho, Minkyoung and Zheng, Boyuan and Gao, Kunxiao and Cao, Yulong and Mao, Z Morley and others},
  journal={arXiv preprint arXiv:2607.17454},
  year={2026}
}

@inproceedings{das2018embodied,
  title     = {Embodied Question Answering},
  author    = {Das, Abhishek and Datta, Samyak and Gkioxari, Georgia and Lee, Stefan and Parikh, Devi and Batra, Dhruv},
  booktitle = {Proceedings of the IEEE Conference on Computer Vision and Pattern Recognition},
  pages     = {1--10},
  year      = {2018}
}

@inproceedings{majumdar2024openeqa,
  title={Openeqa: Embodied question answering in the era of foundation models},
  author={Majumdar, Arjun and Ajay, Anurag and Zhang, Xiaohan and Putta, Pranav and Yenamandra, Sriram and Henaff, Mikael and Silwal, Sneha and Mcvay, Paul and Maksymets, Oleksandr and Arnaud, Sergio and others},
  booktitle={2024 IEEE/CVF Conference on Computer Vision and Pattern Recognition (CVPR)},
  pages={16488--16498},
  year={2024},
  organization={IEEE}
}

@article{ren2024explore,
  title={Explore until confident: Efficient exploration for embodied question answering},
  author={Ren, Allen Z and Clark, Jaden and Dixit, Anushri and Itkina, Masha and Majumdar, Anirudha and Sadigh, Dorsa},
  journal={arXiv preprint arXiv:2403.15941},
  year={2024}
}

@article{zhou2026physvlm,
  title={Physvlm-avr: Active visual reasoning for multimodal large language models in physical environments},
  author={Zhou, Weijie and Xiong, Xuantang and Peng, Yi and Tao, Manli and Zhao, Chaoyang and Dong, Honghui and Tang, Ming and Wang, Jinqiao},
  journal={Advances in Neural Information Processing Systems},
  volume={38},
  pages={3873--3906},
  year={2026}
}

@article{hong2026esibench,
  title={ESI-Bench: Towards Embodied Spatial Intelligence that Closes the Perception-Action Loop},
  author={Hong, Yining and Liu, Jiageng and Yin, Han and Li, Manling and Guibas, Leonidas and Fei-Fei, Li and Wu, Jiajun and Choi, Yejin},
  journal={arXiv preprint arXiv:2605.18746},
  year={2026}
}

@article{sun2026omegaeva,
  title={$\omega $-EVA: Envision, Verify, and Act with Latent Interactive World Models},
  author={Sun, Zhenguo and Sun, Yu and Huang, Hande and Knoll, Alois},
  journal={arXiv preprint arXiv:2606.09457},
  year={2026}
}

@article{cobbe2021training,
  title={Training verifiers to solve math word problems},
  author={Cobbe, Karl and Kosaraju, Vineet and Bavarian, Mohammad and Chen, Mark and Jun, Heewoo and Kaiser, Lukasz and Plappert, Matthias and Tworek, Jerry and Hilton, Jacob and Nakano, Reiichiro and others},
  journal={arXiv preprint arXiv:2110.14168},
  year={2021}
}

@article{yao2023tree,
  title={Tree of thoughts: Deliberate problem solving with large language models},
  author={Yao, Shunyu and Yu, Dian and Zhao, Jeffrey and Shafran, Izhak and Griffiths, Tom and Cao, Yuan and Narasimhan, Karthik},
  journal={Advances in neural information processing systems},
  volume={36},
  pages={11809--11822},
  year={2023}
}

@inproceedings{snell2025scaling,
  title={Scaling {LLM} test-time compute optimally can be more effective than scaling parameters for reasoning},
  author={Snell, Charlie and Lee, Jaehoon and Xu, Kelvin and Kumar, Aviral},
  booktitle={International Conference on Learning Representations},
  volume={2025},
  pages={10131--10165},
  year={2025}
}

@inproceedings{chen2024spatialvlm,
  title     = {{SpatialVLM}: Endowing Vision-Language Models with Spatial Reasoning Capabilities},
  author    = {Chen, Boyuan and Xu, Zhuo and Kirmani, Sean and Ichter, Brian and Sadigh, Dorsa and Guibas, Leonidas and Xia, Fei},
  booktitle = {Proceedings of the IEEE/CVF Conference on Computer Vision and Pattern Recognition},
  pages     = {14455--14465},
  year      = {2024}
}

@inproceedings{cheng2024spatialrgpt,
  title     = {{SpatialRGPT}: Grounded Spatial Reasoning in Vision-Language Models},
  author    = {Cheng, An-Chieh and Yin, Hongxu and Fu, Yang and Guo, Qiushan and Yang, Ruihan and Kautz, Jan and Wang, Xiaolong and Liu, Sifei},
  booktitle = {Advances in Neural Information Processing Systems},
  volume    = {37},
  pages     = {135062--135093},
  year      = {2024}
}

@inproceedings{wu2026spatialmllm,
  title     = {{Spatial-MLLM}: Boosting {MLLM} Capabilities in Visual-Based Spatial Intelligence},
  author    = {Wu, Diankun and Liu, Fangfu and Hung, Yi-Hsin and Duan, Yueqi},
  booktitle = {Advances in Neural Information Processing Systems},
  volume    = {38},
  pages     = {13569--13597},
  year      = {2025}
}

@inproceedings{fan2026vlm3r,
  title     = {{VLM-3R}: Vision-Language Models Augmented with Instruction-Aligned {3D} Reconstruction},
  author    = {Fan, Zhiwen and Zhang, Jian and Li, Renjie and Zhang, Junge and Chen, Runjin and Hu, Hezhen and Wang, Kevin and Wang, Peihao and Qu, Huaizhi and Zhou, Shijie and others},
  booktitle = {Proceedings of the IEEE/CVF Conference on Computer Vision and Pattern Recognition},
  pages     = {31054--31065},
  year      = {2026}
}

@article{yang2025embodiedbench,
  title={Embodiedbench: Comprehensive benchmarking multi-modal large language models for vision-driven embodied agents},
  author={Yang, Rui and Chen, Hanyang and Zhang, Junyu and Zhao, Mark and Qian, Cheng and Wang, Kangrui and Wang, Qineng and Koripella, Teja Venkat and Movahedi, Marziyeh and Li, Manling and others},
  journal={arXiv preprint arXiv:2502.09560},
  year={2025}
}

@article{yao2022react,
  title={React: Synergizing reasoning and acting in language models},
  author={Yao, Shunyu and Zhao, Jeffrey and Yu, Dian and Du, Nan and Shafran, Izhak and Narasimhan, Karthik and Cao, Yuan},
  journal={arXiv preprint arXiv:2210.03629},
  year={2022}
}

@article{schick2023toolformer,
  title={Toolformer: Language models can teach themselves to use tools},
  author={Schick, Timo and Dwivedi-Yu, Jane and Dess{\`\i}, Roberto and Raileanu, Roberta and Lomeli, Maria and Hambro, Eric and Zettlemoyer, Luke and Cancedda, Nicola and Scialom, Thomas},
  journal={Advances in neural information processing systems},
  volume={36},
  pages={68539--68551},
  year={2023}
}

@article{wu2023autogen,
  title={Autogen: Enabling next-gen llm applications via multi-agent conversation},
  author={Wu, Qingyun and Bansal, Gagan and Zhang, Jieyu and Wu, Yiran and Li, Beibin and Zhu, Erkang and Jiang, Li and Zhang, Xiaoyun and Zhang, Shaokun and Liu, Jiale and others},
  journal={arXiv preprint arXiv:2308.08155},
  year={2023}
}

@inproceedings{liang2023code,
  title={Code as policies: Language model programs for embodied control},
  author={Liang, Jacky and Huang, Wenlong and Xia, Fei and Xu, Peng and Hausman, Karol and Ichter, Brian and Florence, Pete and Zeng, Andy},
  booktitle={2023 IEEE International conference on robotics and automation (ICRA)},
  pages={9493--9500},
  year={2023},
  organization={IEEE}
}

@article{zhang2026harness,
  title={Harness VLA: Steering Frozen VLAs into Reliable Manipulation Primitives via Memory-Guided Agents},
  author={Zhang, Yixian and Zhang, Huanming and Gao, Feng and Li, Xiao and Liu, Zhihao and Zhu, Chunyang and Qiu, Jiaxing and Yan, Yuchen and Liu, Jiyuan and Tang, Wenhao and others},
  journal={arXiv preprint arXiv:2607.08448},
  year={2026}
}

@article{zhou2025libero,
  title={Libero-pro: Towards robust and fair evaluation of vision-language-action models beyond memorization},
  author={Zhou, Xueyang and Xu, Yangming and Tie, Guiyao and Chen, Yongchao and Zhang, Guowen and Chu, Duanfeng and Zhou, Pan and Sun, Lichao},
  journal={arXiv preprint arXiv:2510.03827},
  year={2025}
}

@article{wang2024executable,
  title={Executable code actions elicit better llm agents},
  author={Wang, Xingyao and Chen, Yangyi and Yuan, Lifan and Zhang, Yizhe and Li, Yunzhu and Peng, Hao and Ji, Heng},
  journal={arXiv preprint arXiv:2402.01030},
  year={2024}
}

@article{yang2024swe,
  title={Swe-agent: Agent-computer interfaces enable automated software engineering},
  author={Yang, John and Jimenez, Carlos and Wettig, Alexander and Lieret, Kilian and Yao, Shunyu and Narasimhan, Karthik and Press, Ofir},
  journal={Advances in Neural Information Processing Systems},
  volume={37},
  pages={50528--50652},
  year={2024}
}

@inproceedings{wang2025openhands,
  title={Openhands: An open platform for ai software developers as generalist agents},
  author={Wang, Xingyao and Li, Boxuan and Song, Yufan and Xu, Frank F and Tang, Xiangru and Zhuge, Mingchen and Pan, Jiayi and Song, Yueqi and Li, Bowen and Singh, Jaskirat and others},
  booktitle={International Conference on Learning Representations},
  volume={2025},
  pages={65882--65919},
  year={2025}
}

@article{lee2026meta,
  title={Meta-harness: End-to-end optimization of model harnesses},
  author={Lee, Yoonho and Nair, Roshen and Zhang, Qizheng and Lee, Kangwook and Khattab, Omar and Finn, Chelsea},
  journal={arXiv preprint arXiv:2603.28052},
  year={2026}
}

@article{fu2026cap,
  title={CaP-X: A framework for benchmarking and improving coding agents for robot manipulation},
  author={Fu, Letian and Yu, Justin and El-Refai, Karim and Kou, Ethan and Xue, Haoru and Huang, Huang and Xiao, Wenli and Wang, Guanzhi and Niu, Dantong and Li, Fei-Fei and others},
  journal={arXiv preprint arXiv:2603.22435},
  year={2026}
}

@article{lu2026aspire,
  title={ASPIRE: Agentic/Skills Discovery for Robotics},
  author={Lu, Runyu and Wu, Yubo and Kou, Ethan and Fu, Letian and Xiao, Wenli and Mandlekar, Ajay and Xu, Yinzhen and Shi, Guanya and Goldberg, Ken and Chen, Ang and others},
  journal={arXiv preprint arXiv:2607.00272},
  year={2026}
}

@article{lei2026towards,
  title={Towards Long-horizon Embodied Agents with Tool-Aligned Vision-Language-Action Models},
  author={Lei, Zixing and Liu, Changxing and Xiong, Yichen and Xiong, Minhao and Ding, Yuanzhuo and Zhang, Zhipeng and Li, Weixin and Chen, Siheng},
  journal={arXiv preprint arXiv:2605.13119},
  year={2026}
}

@article{kim2024openvla,
  title={Openvla: An open-source vision-language-action model},
  author={Kim, Moo Jin and Pertsch, Karl and Karamcheti, Siddharth and Xiao, Ted and Balakrishna, Ashwin and Nair, Suraj and Rafailov, Rafael and Foster, Ethan and Lam, Grace and Sanketi, Pannag and others},
  journal={arXiv preprint arXiv:2406.09246},
  year={2024}
}

@article{black2024pi_0,
  title={$\pi_0 $: A Vision-Language-Action Flow Model for General Robot Control},
  author={Black, Kevin and Brown, Noah and Driess, Danny and Esmail, Adnan and Equi, Michael and Finn, Chelsea and Fusai, Niccolo and Groom, Lachy and Hausman, Karol and Ichter, Brian and others},
  journal={arXiv preprint arXiv:2410.24164},
  year={2024}
}

@article{intelligence2025pi_,
  title={$\pi_{0.5}$ : a Vision-Language-Action Model with Open-World Generalization},
  author={Intelligence, Physical and Black, Kevin and Brown, Noah and Darpinian, James and Dhabalia, Karan and Driess, Danny and Esmail, Adnan and Equi, Michael and Finn, Chelsea and Fusai, Niccolo and others},
  journal={arXiv preprint arXiv:2504.16054},
  year={2025}
}

@article{lee2025molmoact,
  title={Molmoact: Action reasoning models that can reason in space},
  author={Lee, Jason and Duan, Jiafei and Fang, Haoquan and Deng, Yuquan and Liu, Shuo and Li, Boyang and Fang, Bohan and Zhang, Jieyu and Wang, Yi Ru and Lee, Sangho and others},
  journal={arXiv preprint arXiv:2508.07917},
  year={2025}
}

@article{hung2025nora,
  title={Nora: A small open-sourced generalist vision language action model for embodied tasks},
  author={Hung, Chia-Yu and Sun, Qi and Hong, Pengfei and Zadeh, Amir and Li, Chuan and Tan, U and Majumder, Navonil and Poria, Soujanya and others},
  journal={arXiv preprint arXiv:2504.19854},
  year={2025}
}

@inproceedings{zheng2026x,
  title={X-vla: Soft-prompted transformer as scalable cross-embodiment vision-language-action model},
  author={Zheng, Jinliang and Li, Jianxiong and Wang, Zhihao and Liu, Dongxiu and Kang, Xirui and Feng, Yuchun and Zheng, Yinan and Zou, Jiayin and Chen, Yilun and Zeng, Jia and others},
  booktitle={International Conference on Learning Representations},
  volume={2026},
  pages={60580--60606},
  year={2026}
}

@article{sun2026atomvla,
  title={Atomvla: Scalable post-training for robotic manipulation via predictive latent world models},
  author={Sun, Xiaoquan and Xu, Zetian and Cao, Chen and Liu, Zonghe and Sun, Yihan and Pang, Jingrui and Zhang, Ruijian and Yang, Zhen and Pang, Kang and He, Dingxin and others},
  journal={arXiv preprint arXiv:2603.08519},
  year={2026}
}

@article{yang2025steering,
  title={Steering vision-language-action models as anti-exploration: A test-time scaling approach},
  author={Yang, Siyuan and Zhang, Yang and He, Haoran and Pan, Ling and Li, Xiu and Bai, Chenjia and Li, Xuelong},
  journal={arXiv preprint arXiv:2512.02834},
  year={2025}
}

@inproceedings{ye2026tts,
  title={E-TTS: A New Embodied Test-Time Scaling Framework for Robotic Manipulation},
  author={Ye, Wen and Li, Peiyan and Yuan, Tingyu and Xu, Yuan and Wu, Xiangnan and Zhao, Chaoyang and Liu, Jing and Liu, Nianfeng and Huang, Yan and Wang, Liang},
  booktitle={European Conference on Computer Vision},
  pages={395--412},
  year={2026},
  organization={Springer}
}

@article{chen2021history,
  title={History aware multimodal transformer for vision-and-language navigation},
  author={Chen, Shizhe and Guhur, Pierre-Louis and Schmid, Cordelia and Laptev, Ivan},
  journal={Advances in neural information processing systems},
  volume={34},
  pages={5834--5847},
  year={2021}
}

\end{document}